\pdfoutput=1 
\documentclass[10pt,twocolumn,letterpaper]{article}
\newcommand{\bfp}{{\bf p}}

\newcommand{\bfel}{{\boldsymbol\ell}}

\usepackage{cvpr}
\usepackage{times}
\usepackage{epsfig}
\usepackage{amsmath}
\usepackage{amssymb}
\usepackage{array}
\usepackage{amsthm}
\usepackage{comment}
\usepackage{multirow}
\usepackage{overpic}
\usepackage{enumitem}
\usepackage{graphicx}
\usepackage{float}
\usepackage{amsbsy}
\usepackage{makecell}

\usepackage[pagebackref=true,breaklinks=true,letterpaper=true,colorlinks,bookmarks=false]{hyperref}

\cvprfinalcopy 
\def\cvprPaperID{1931} % *** Enter the CVPR Paper ID here

\ifcvprfinal\fi
\begin{document}

%%%%%%%%% TITLE
\title{Learning to Separate Multiple Illuminants in a Single Image}

\author{Zhuo Hui,$^{1}$ Ayan Chakrabarti,$^{2}$ Kalyan Sunkavalli,$^{3}$ and Aswin C.\ Sankaranarayanan$^{1}$\\
	$^{1}$Carnegie Mellon University \hspace{10mm}  $^{2}$Washington University in St. Louis \hspace{10mm} $^{3}$Adobe Research \hspace{3mm}
}
\maketitle

\begin{abstract}
We present a method to separate a single image captured under two illuminants, with different spectra, into the two images corresponding to the appearance of the scene under each individual illuminant. We do this by training a deep neural network to predict the per-pixel reflectance chromaticity of the scene, which we use in a physics-based image separation framework to produce the desired two output images. We design our reflectance chromaticity network and loss functions by incorporating intuitions from the physics of image formation. We show that this leads to significantly better performance than other single image techniques and even approaches the quality of the prior work that require additional images.  
\end{abstract}

\section{Introduction} \label{sec:intro}

Natural environments are often lit by multiple light sources with different illuminant spectra. 
Depending on scene geometry and material properties, each of these lights causes different light transport effects like color casts, shading, shadows, specularities, etc.
An image of the scene combines the effects from the different lights present, and is a superposition of the images that would have been captured under each individual light.
We seek to invert this superposition, i.e., separate a single image observed under two light sources, with different spectra, into two images, each corresponding to the appearance of the scene under one light source alone.
Such a decomposition can give users the ability to edit and relight photographs, as well as provide information useful for photometric analysis.

However, the appearance of a surface depends not only on the properties of the light sources, but also on its geometry and material properties. 
When all of these quantities are unknown, disentangling them is a significantly ill-posed problem.
Thus, past efforts to achieve such separation have relied heavily on extensive manual annotation~\cite{bousseau2009user,bonneel2014interactive,boyadzhiev2012user} or access to calibrated scene and lighting information~\cite{Debevec12:LightStage,debevec2008rendering}.
More recently,  Hui et al.\ \cite{hui2016white,Hui_2018_CVPR} demonstrate that the lighting separation problem can be reliably solved if one additionally knows the reflectance chromaticity of all surface points --- which they recover by capturing a second image of the same scene under flash lighting. 
Given that the flash image is used in their processing pipeline only for estimating the reflectance chromaticity, could we computationally estimate the reflectance chromaticity from a \emph{single image}, thereby avoiding the need to capture a flash photograph all together?
This would greatly enhance the applicability of the method especially for scenarios where it is challenging to sufficiently illuminate every pixel with the flash; for example, when the flash is not strong enough, the scene is large, or the ambient light sources are too strong.

\begin{figure}[!t]
	\centering
	\setlength{\tabcolsep}{1pt}
	\begin{tabular}{ccc}
		\includegraphics[clip,width=.32\linewidth]{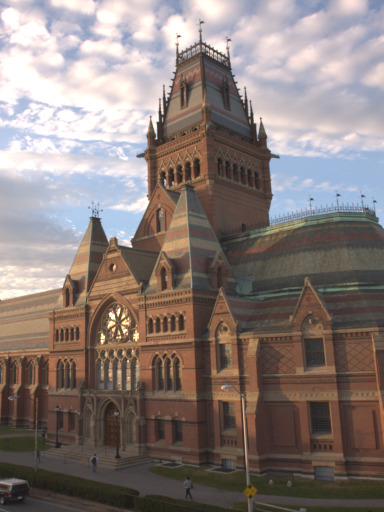} & 
		\includegraphics[clip,width=.32\linewidth]{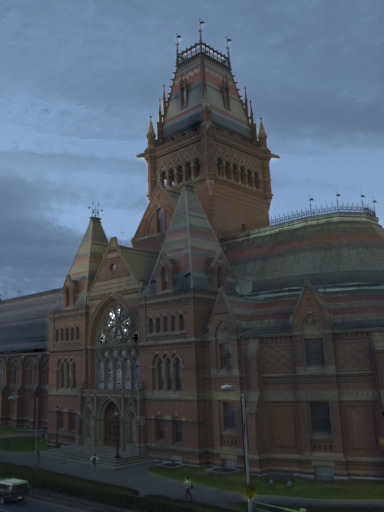} &
		\includegraphics[clip,width=.32\linewidth]{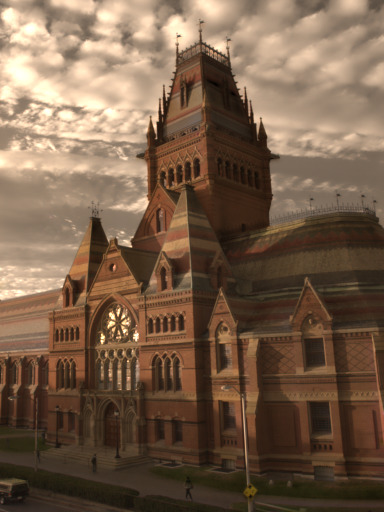} \\
		(a) Input image & \multicolumn{2}{c}{(b) Output separated images} \\
	\end{tabular}
	\caption{Our method separates a single image (a) captured under two illuminants with different spectra (sun and sky illumination here) into two images corresponding to the appearance of the scene under the individual lights. Note that we are able to accurately preserve the shading and shadows for each light.}
	\label{fig:teaser}
\end{figure}

Our work is also motivated by the success of deep convolutional neural networks for solving closely related problems like intrinsic decompositions~\cite{li2018cgintrinsics,zhou2015learning}, and reflectance estimation~\cite{tang2012deep,rematas2016deep,li2018materials}; hence, we propose training a deep convolution neural network to perform this separation.
However, we find that standard architectures, trained only with the respect to the quality of the final separated images, are unable to learn to effectively perform the separation.
Therefore, we guide the design of our network using a physics-based analysis of the task~\cite{Hui_2018_CVPR} to match the expected sequence of inference steps and intermediate outputs --- reflectance chromaticities, shading chromaticities, separated shading maps, and final separated images.
In addition to ensuring that our architecture has the ability to express these required computations, this decomposition also allows us to provide supervision to intermediate layers in our network, which proves crucial to successful training.

We train our network on two existing datasets: the synthetic database of Li et al.~\cite{li2018cgintrinsics}, and the set of real flash/no-flash pairs collected by Aksoy et al.~\cite{flashambient} --- using a variant of Hui et al.'s algorithm~\cite{Hui_2018_CVPR} to compute ground-truth values. Once trained, we find that our approach is able to successfully solve this ill-posed problem and produce high-quality lighting decompositions that, as can be seen in Figure~\ref{fig:teaser}, capture complex shading and shadows. In fact, our network is able to match, and in specific instances outperform, the quality of results from Hui et al.'s two-image method~\cite{Hui_2018_CVPR}, despite needing only a single image as input.

\section{Related Work}\label{sec:prior}

Estimating illumination and scene geometry from a single image is a highly ill-posed problem. 
Previous work has focused on specific subsets of this problem; we discuss previous works on illumination analysis as well as prior intrinsic image decomposition methods that aim to jointly estimate illumination and surface reflectance.

\paragraph{Illumination estimation.} 
Estimating the ambient illumination from a single photograph has been a long-standing goal in computer vision and computer graphics.
A number of past techniques have studied color constancy~\cite{Gijsenij2011survey} --- the problem of removing the color casts of ambient illumination.
One popular solution is to model the scene with single dominant light source~\cite{finlayson1993diagonal,finlayson1993enhancing,gehler2008bayesian}.
To deal with mixtures of illuminants in a scene, previous works \cite{ebner2004color,gijsenij2012color,riess2011illuminant} typically characterize each local region with a different but single light source.
However, these approaches do not generalize well to scenes where multiple light sources mix smoothly.
To address this, Boyadzhiev et al.~\cite{boyadzhiev2013user} utilize user scribbles to indicate scene attributes such as white surfaces and constant lighting regions.
Hsu et al.~\cite{hsu2008light} propose a method to address mixtures of two light sources in the scene; however, they require precise knowledge of the color of each illuminant.
Prinet et al.~\cite{prinet2013illuminant} resolve the color chromaticity of two light sources from a sequence of images by leveraging the consistency of the reflectance of the scene.
Sunkavalli et al.~\cite{sunkavalli08color} demonstrate this (and image separation) for time-lapse sequences of outdoor scenes.

In parallel, many techniques have been developed to explicitly model the illumination of the scene, rather than removing the color of the illuminants.
Lalonde et al.~\cite{lalonde2010sun} propose the parametric model to characterize the sky and sun for the outdoor photographs.
Hold-Geoffroy et al.~\cite{hold2017deep} extend the idea to model the outdoor illumination by incorporating a deep neutral network.
Gardner et al.~\cite{gardner2017indoor} similarly train a deep neural network to recover indoor illumination from a single LDR photograph.
In contrast, our method does not explicitly model the illuminants, but directly regresses the single-illuminant images.

\paragraph{Intrinsic image decomposition.}
Intrinsic image decomposition methods seek to separate a single image into a product of reflectance and illumination layers. 
This problem is commonly solved by assuming that the reflectance of the scene is piece-wise constant while the illumination varies smoothly~\cite{Barrow78:Intrinsic}.
Several approaches build on this by further imposing priors on non-local reflectance~\cite{zhao2012closed,shen2013intrinsic,bell2014intrinsic}, or on the consistency of reflectance for image sequences captured with varying illumination~\cite{Laffont:ICCV15,hauagge2013photometric,Hui_2017_ICCV}. 
A common assumption in intrinsic image methods is that the scene is lit by a single dominant illuminant.
% in practice, this allows them to assume that the shading layer is monochromatic (and the color of the illuminant is either absorbed into the reflectance or represented by a single global parameter).
%
This does not generalize to real world scenes that are often illuminated with multiple light sources with different spectra. 
Recent methods have proposed using deep neural networks, trained with large amounts of data, to address this problem~\cite{zhou2015learning,li2018cgintrinsics,li2018learning}.
While effective, these techniques also focus on scenes illuminated with a single light source.
Barron and Malik~\cite{barron2012color,barron2015shape} resolve this by incorporating a global lighting model with hand-crafted priors. 
While this lighting model works well for single objects, it is unable to capture high-frequency spatial information, like shadows that are often present in real scenes.
In comparison, our technique generalizes well to complex scenes lit with mixtures of multiple light sources.  
In addition, as opposed to predicting the reflectance of the scene, our method only requires predicting its \textit{chromaticity}, which is an easier problem to solve.
%
%Recently, Hui et al.~\cite{hui2016white,Hui_2018_CVPR} resolve this by taking an extra photograph using flash light.
%%
%While this technique produces high quality results, they are inherently limited by the use of flash since it is inapplicable to weak flash and motion between the capture image pair.
%%
%In contrast, we only use a single photograph and directly regress the separated image without relying on the flash light, making it well-suited for portable device and smart phones.

\section{Problem Statement}\label{sec:overview}
%!TEX root = main.tex
\begin{figure*}
	{
		\centering
		\setlength{\tabcolsep}{1pt}
		\begin{tabular}{c}		
			
			\includegraphics[width=\textwidth]{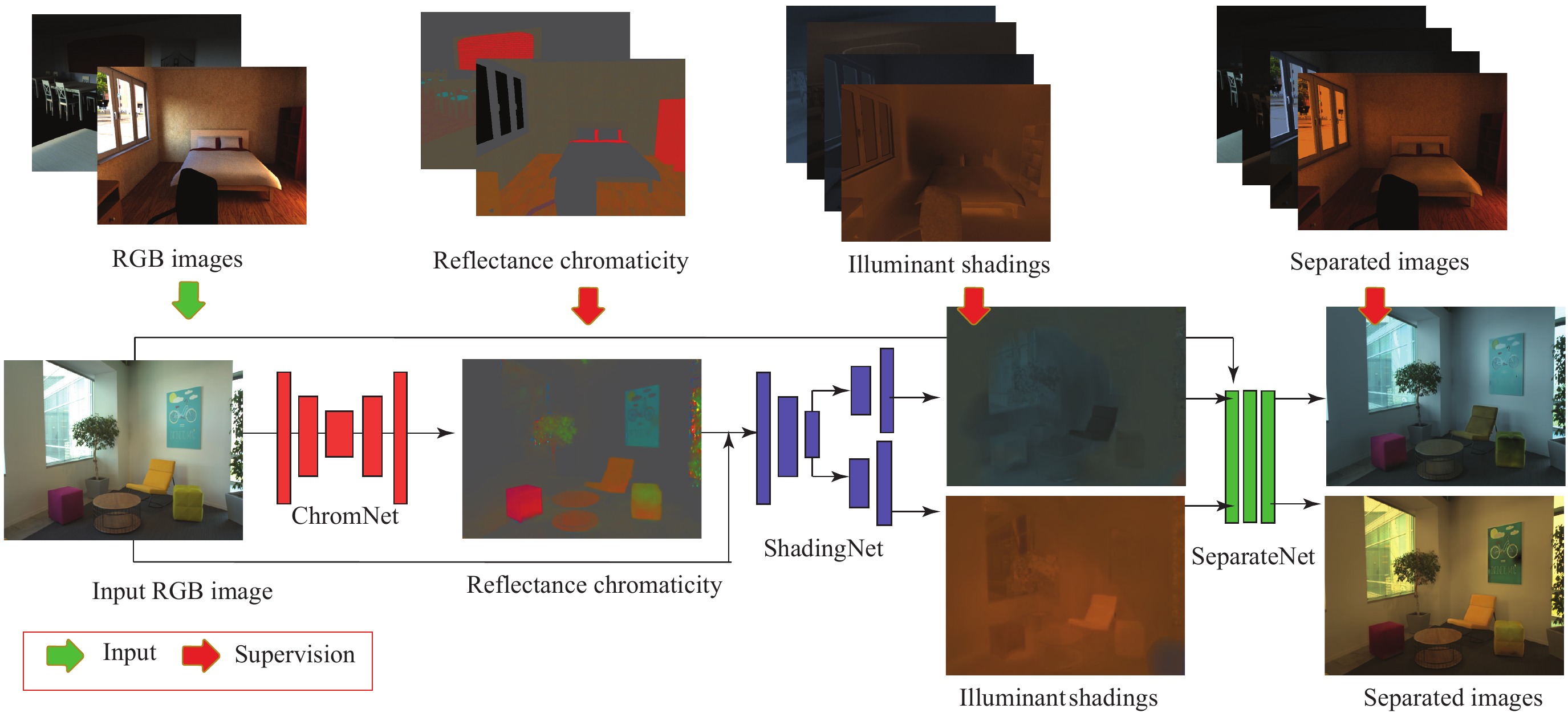}\\
			
		\end{tabular}
		
		\caption{Given a single image under the mixture of lighting, our method automatically produces the images lit by each of these illuminants.  We train a cascade of three sub-networks with three specific tasks. First, we estimate the reflectance color chromaticity of the scene via  ChromNet. Given this estimation, we concatenate it with the input RGB image and feed them into ShadingNet to predict the illuminant shadings. We append these to the input image and pass it to SeparateNet to produce the output. During training, we supervise the reflectance chromaticity, illuminant shadings and the separated images. }
		%This allows us to incorporate physics-based constraints and  regularize the separation process.}
		\label{fig:overview}
		
	}
\end{figure*}

Our objective is to take as input, a single photograph of a scene lit by a mixture of two illuminants, and estimate the images lit by each single light source. 
%
%This is a severely ill-posed problem and we propose solving it using deep neural networks.
%
In this section, we set up the image formation model and describe the physical priors we impose to supervise the intermediate results produced by the network.

%

%\subsection{Problem setup and image formation}
%
We adopt the image formation model from Hui et al.~\cite{Hui_2018_CVPR} by assuming that the scene is Lambertian and is imaged by a three-channel color camera.
However, instead of modeling infinite-dimensional spectra using subspaces, we assume that the camera color response is narrow-band, allowing us to characterize both the light source and albedo in RGB.
That is, the intensity observed at a pixel $\bfp$ in a single photograph $I$ is given by:
\begin{equation}
I^c(\bfp) = R^c(\bfp)\sum_{i=1}^N \lambda_i(\bfp)\ \ell_i^c, \quad \textrm{for } c \in \{r, g, b\},
\label{eq:im_n_ini}
\end{equation}
where $R(\bfp) = [ R^r(\bfp),  R^g(\bfp),  R^b(\bfp) ]$ is the three-color albedo. 
In our work, we focus on the scenes that are lit by $N = 2$ light sources and we denote the light chromaticities as $\{ \bfel_1, \bfel_2\}$.
Note that $\bfel_i = [ \ell_i^r, \ell_i^g, \ell_i^b ] \in {\mathbb R}^3$ with $\sum_{c} \ell_i^c = 1$. 
Similar to Hui et al.~\cite{Hui_2018_CVPR}, we assume that the light source chromaticities are unique, i.e., $ \bfel_1 \ne \bfel_2$.
The term $\lambda_i(\bfp)$ is the shading observed at pixel $\bfp$ due to the $i$-th light source multiplied by the light-source brightness.
Given the fact that two sources with the same color are clustered together, the shading term $\lambda_i(\bfp)$ has a complex dependency on the lighting geometry and does not have a simple analytical form.
Our goal is to compute the separated images corresponding to the each light source $k$ as:
\begin{equation}
 \widehat{I}^c_{\textrm{sep}, k }(\bfp) = R^c(\bfp)\  \lambda_k(\bfp) \  {\ell_k^c}.
\end{equation}

%To separate the photograph into two images is a severely under-constrained problem. 
%
To solve this, Hui et al.~\cite{Hui_2018_CVPR} capture an additional image under flash illumination to directly compute reflectance color chromaticities for each pixel. They use these to disentangle reflectance from illumination shading,
and solve for the color of each light source as well as the per-pixel contribution of each illuminant.
%
%Inspired by their work, we propose a two-stage network architecture (as shown in Figure~\ref{fig:teaser}).
We provide a quick summary of the key steps of their computational pipeline below, and refer readers to their paper for more details.%, adapted to the RGB color model.

{{\flushleft \textit{Step 1 --- Flash to reflectance chromaticity.}} Given the flash color, the pure flash photograph enables us to estimate the reflectance chromaticity, $\alpha^c(\bfp)$, that is defined as:
%Following their work, we describe the image separation algorithm based on the assumed RGB model in (\ref{eq:im_n_ini}).
%
%Specifically, we first predict the reflectance color chromaticity defined as:
\begin{equation}
\alpha^c(\bfp) = \frac{R^c(\bfp)}{\sum_{\tilde{c}} R^{\tilde{c}}(\bfp)}.
\label{eq:rho_chrom}
\end{equation} 

{{\flushleft \textit{Step 2 --- Estimate shading chromaticity.}} The reflectance chromaticity $\alpha^c(\bfp)$ can be used to remove the contribution from albedo as follows.
We define $\beta^c(\bfp) =  {I^c(\bfp)}/{\alpha^c(\bfp)}$
and normalize it to obtain shading chromaticities as:
\begin{equation}
\gamma^c(\bfp) = \frac{\beta^c(\bfp)}{\sum_{\tilde{c}}\beta^{\tilde{c}}(\bfp)} =  \frac{\sum_{i} \lambda_i(\bfp)\ \ell_i^c}{\sum_{i} \lambda_{i}(\bfp)} = \sum_{i} z_i(\bfp) \ \ell_i^c,
\label{eq:s_img}
\end{equation}
where $z_i(\bfp) = \lambda_i(\bfp) / \left(\sum_{\tilde{i}} \lambda_{\tilde{i}}(\bfp)\right)$ is relative shading.
%
%This term captures the fraction of the shading at a scene pixel that comes from one light source, relative to all the light sources, hence the term \emph{relative} shading. 
%
%Note that,  $z_i(\bfp) \ge 0$ since $\lambda_k(\bfp)$, the shading associated with each light source at a pixel, are non-negative.
%
%Also note that, $\sum_i z_i(\bfp) = 1$ by construction.
%
%The shading chromaticity $\gamma$ is key for source separation with respect to the illuminant colors. 

{{\flushleft \textit{Step 3 --- Estimate relative shading.}}
The histogram of shading chromaticities across the image can be fit to a multi-illuminant model (see \cite{Hui_2018_CVPR} for details) to estimate the illuminant shadings $S_i^c$ for each light source, defined as:
\begin{equation}
S_i^c(\bfp) = z_i(\bfp) \ell_i^c.
\label{eq:illuminant_shadings}
\end{equation}
The separated images can then be recovered as:
\begin{equation}
\widehat{I}^c_{\textrm{sep}, k }(\bfp) = I^c(\bfp) \frac{S_k^c(\bfp)}{\sum_{i = 1}^N S_i^c(\bfp)}.
\label{eq:img_sep_final}
\end{equation}

In this paper, we design our network by mimicking the steps in the derivation above, but each processing element is replaced with deep networks as shown in Figure~\ref{fig:overview}.
In particular, we utilize three sub-networks --- ChromNet, ShadingNet and SeparateNet --- to estimate the reflectance chromaticity, illuminant shadings and separated images, respectively. 
ChromNet predicts the values of reflectance chromaticity $\alpha$, defined in (\ref{eq:rho_chrom}), with its input being the RGB image that we seek to separate.
ShadingNet takes in as the output of ChromeNet concatenated with the input RGB image to regress the illuminant shadings in (\ref{eq:illuminant_shadings}).
Finally, SeparateNet gathers the estimated illuminant shadings as well as the input RGB image to estimate the separated images.

\section{Learning Illuminant Separation}
We now dicuss our proposed method for decomposing an input photo into images lit by individual illuminants, including how we generate training data with ground truth scene annotations (for reflectance and shading), and how we design our proposed deep neural network for this problem. 
%Given training data with intermediate scene values (reflectance chromaticity and relative shadings) and final separated images, we next describe our method for decomposing an input photo into images lit by individual illuminants.

\subsection{Generating the training dataset}
We utilize the databases of CGIntrinsics~\cite{li2018cgintrinsics} and Flash/No-Flash~\cite{flashambient} to produce images with (approximate) ground truth reflectance chromaticity, illuminant shadings and separated images.
Figure~\ref{fig:train_samples} shows training data examples from each dataset.

The CGIintrinsics dataset consists of $20160$ rendered scenes from SUNCG~\cite{song2016ssc}. It provides the ground truth reflectance, and hence, the reflectance chromaticity.
The shadings chromaticity is then estimated via (\ref{eq:s_img}).

The Flash/No-flash dataset consists of $2775$ image pairs. We estimate the reflectance chromaticity as the color chromaticity of the pure flash image, which is the difference between the flash and the no-flash photograph.
%
%In particular, the flash image in the database is illuminated only by the flash light, hence, we denote the intensity observed as
%\begin{equation}
%I_{\textrm{pf}}^c(\bfp) = \rho^c(\bfp)  \lambda_f(\bfp)\ \ell_f^c, \quad \textrm{for } c \in \{r, g, b\},
%\label{eq:im_f}
%\end{equation}
%%
%where $\lambda_f(\bfp)$ denotes the shading at $\bfp$ caused by the flash, and the chromaticity of the flash $\bfel_f = [ \ell_f^r, \ell_f^g, \ell_f^b ]$ is assumed to known as $[1 1 1]/3$.
%%
%Note that this allows us to recover the albedo chromaticity upto the unknown flash color but still yields accurate separated images
%%
%Dividing by the flash light chromaticity, $\bfel_f$, gives us the reflectance chromaticity
%%
%\begin{equation}
%\alpha^c(\bfp) = I_{\pf}^c({\bfp}) / \ell_{f}^c = \rho^c(\bfp) \lambda_{f} (\bfp).
%\label{eq:alpha}
%\end{equation}
%
We anecdotally observed that the majority of the scenes in this dataset are only illuminated by a single light source --- which, as such, makes it uninteresting for our application.
To resolve this, we add the flash image back to no-flash image and create photographs illuminated by two light sources.
By changing the color of the flash photograph, we can enhance the amount of training data; this allows us to generate $29060$ input-output pairs, where the input is a photo, and the output is the reflectance chromaticity, a pair of its corresponding illuminant shadings as well as the separated images.

\begin{figure}{
					\small
					\centering
					\setlength{\tabcolsep}{1pt}
					
					\begin{tabular}{ccc}
						\rotatebox{90}{\hspace{0.2cm}CGIntrinsics}		
						\includegraphics[width=0.16\textwidth]{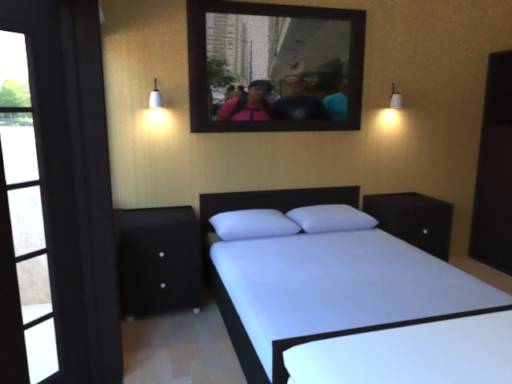}&
						\includegraphics[width=0.16\textwidth]{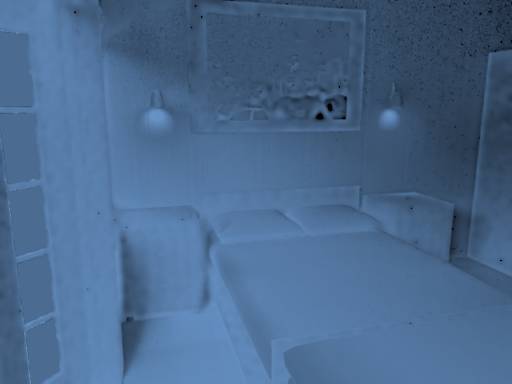}&
						\includegraphics[width=0.16\textwidth]{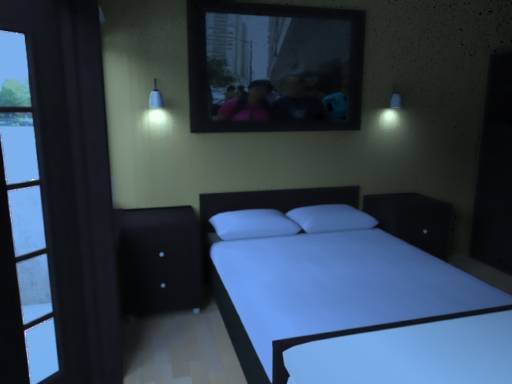}\\
						
						\rotatebox{90}{\hspace{0.2cm}CGIntrinsics}		
						\includegraphics[width=0.16\textwidth]{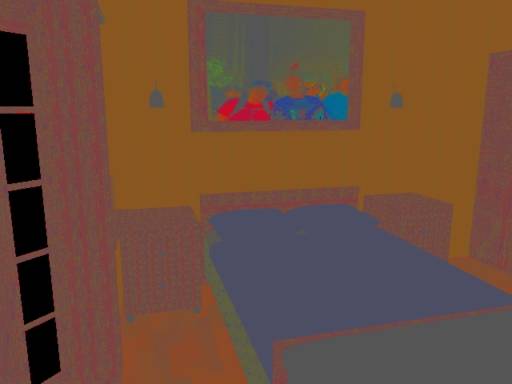}&
						\includegraphics[width=0.16\textwidth]{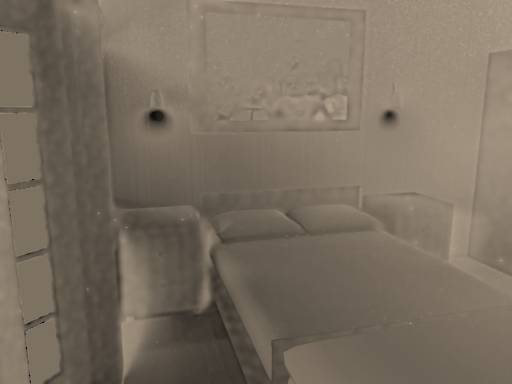}&
						\includegraphics[width=0.16\textwidth]{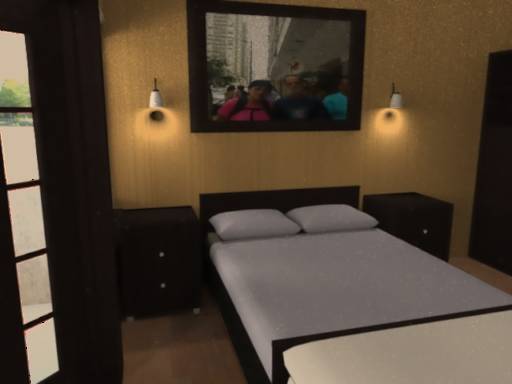}\\			
						
						\rotatebox{90}{\hspace{0.2cm}Flash/No-flash}	
						\includegraphics[width=0.16\textwidth]{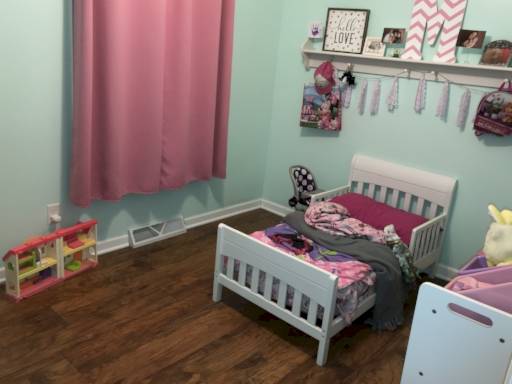}&
						\includegraphics[width=0.16\textwidth]{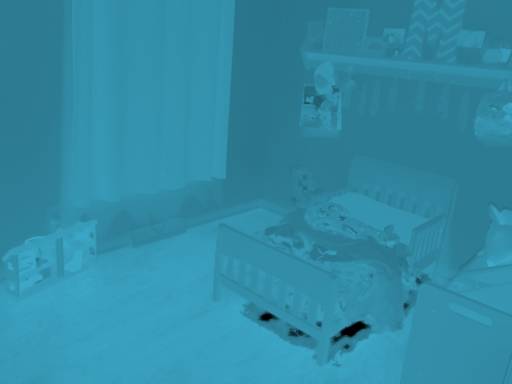}&
						\includegraphics[width=0.16\textwidth]{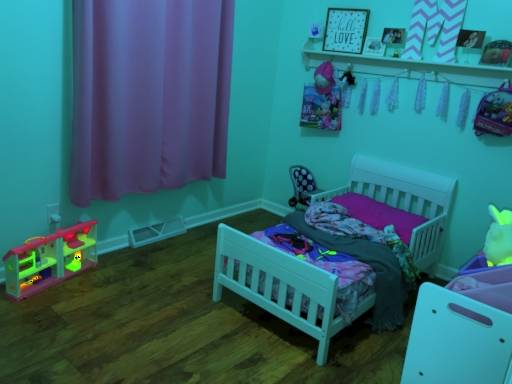}\\
						\rotatebox{90}{\hspace{0.2cm}Flash/No-flash}	
						\includegraphics[width=0.16\textwidth]{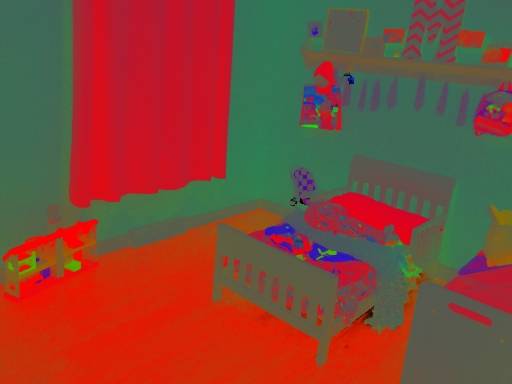}&
						\includegraphics[width=0.16\textwidth]{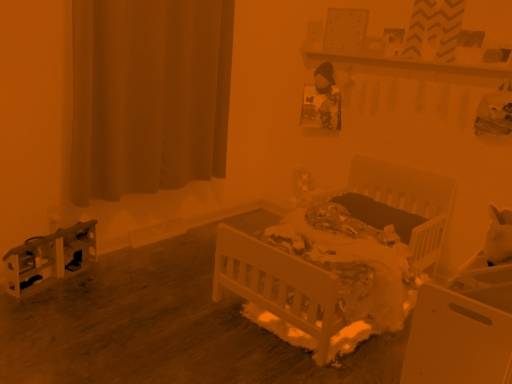}&
						\includegraphics[width=0.16\textwidth]{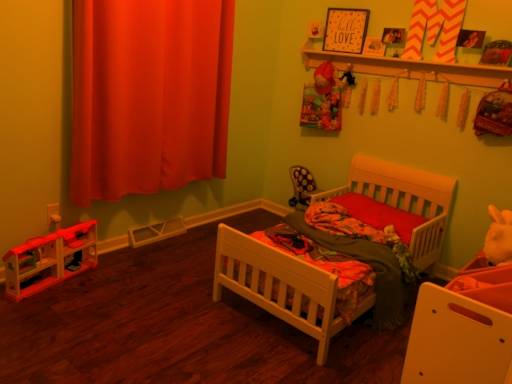}\\			
						(a) Input (Top) / & (b) Illuminant  & (c) Separated \\
						Chromaticity (Bottom) & shadings & images \\

					\end{tabular}
					
					\caption{We showcase each sample of train pairs from CGIntrinsics (top) and Flash/No-flash database (bottom). Given the input photograph (a), we use reflectance chromaticity  together with illuminant shadings (b) and separated images (c) to supervise the output of the network. 
					}
					\label{fig:train_samples}
					
				}
			\end{figure}
			
			%\paragraph{Remarks.}
			%If we consider our goal as a learning problem, the supervised image-to-image translation would be ideal for this learning task.
			%
			%That said, training data in the form of {photograph, separated images} would be desirable.
			%
			%However, we found that naively applying image-to-image translation  always leads to large errors since it is difficult to correct the artifacts only by regularizing the output term.  
			%
			%In Figure~\ref{fig:comp_img2img}, we compare the proposed architecture against the U-Net, which takes a single photograph and predicts two separated images.
			%
			%As can be seen here, the performance of the propose method improves significantly by imposing the physical priors on the intermediate results. 
			%

			%

\subsection{Network architecture}
As shown in Figure~\ref{fig:overview}, we use a deep neural network to match the computation of the separation algorithm in Section~\ref{sec:overview}. 
Specifically, our network consists of three sub-networks that produce the reflectance chromaticity, illuminant shadings, and the separated images respectively.

\paragraph{ChromNet.} We design the first sub-network to explicitly estimate the reflectance chromaticity~(\ref{eq:rho_chrom}) from the input color image.
This essentially requires the network to solve the ill-posed problem of estimating and removing the illumination color cast given only a single photograph.
%
%Note that there is no analytic form solution for this problem from a single photograph.
%
%In that sense, it can be considered as a neural style transfer problem, from a single input under mixture of illumination to the reflectance chromaticity without the light color casts. \KS{weird justification}
%
%
We adopt an architecture similar to that of Johnson et al.~\cite{johnson2016perceptual} to map the input image to a three channel reflectance chromaticity map.\footnote{A detailed description of the construction of each subnetwork is provided in the supplemental material.}

%which consists of a sequence of a $7\times 7$ convolution layer, two $3 \times 3$ stride-$2$ convolution layers, nine residual blocks (with two $3\times 3$ convolution layers each), two up-sampling blocks that each perform nearest neighbor up-sampling by a factor of $2$ followed by a $3\times 3$ convolution, and a final $7\times 7$ convolution layer with tanh activation to produce the reflectance chromaticity estimates. 
%

%

\paragraph{ShadingNet.}

The second sub-network in our framework takes reflectance chromaticity estimates as inputs, and solves for the two illuminant shadings in (\ref{eq:illuminant_shadings}).
From Section~\ref{sec:overview}, we expect the first part of this computation to involve deriving $\gamma$ from the chromaticities and original input, on a purely per-pixel basis as per (\ref{eq:s_img}). However, we found computing the $\gamma$ values explicitly to lead to instability in training, likely since this involves a division. Instead, we produce a general feature map intended to encode the $\gamma$ information (note that we do not require it to exactly correspond to $\gamma$ values) by concatenating the input image with the estimated chromaticities.
Given this feature map, our second sub-network produces the two separated illuminant shading maps. Since this requires global reasoning, we use an architecture similar to the pixel-to-pixel network of Isola et al.~\cite{isola2017image} to incorporate a large receptive field.

\paragraph{SeparateNet.}
Given the illuminant shadings and previously estimated reflectance chromaticity, the last computation step is to produce the separated image. Here, we use a series of pixel-wise layers to express the computation in (\ref{eq:img_sep_final}). Our third sub-network concatenates the two predicted shading maps and the input RGB photograph  into a nine-channel input, and uses three $1\times 1$ convolution layers to produce a six-channel output corresponding to the two final separated RGB images.

Note that the output of our first sub-network---reflectance chromaticity---is sufficient to perform separation using the method of Hui et al.~\cite{Hui_2018_CVPR}. However, training this sub-network based directly on the quality of reflectance chromaticity estimates proves insufficient, because the final separated image quality can degrade differently with different kind of errors in chromaticity estimates. Thus, our goal is to instead train the reflectance chromaticity estimation sub-network to be optimal towards final separation quality. Unfortunately, the separation algorithm in \cite{Hui_2018_CVPR} has non-differentiable processing steps, as well as other computation that produces unstable gradients. Hence, we use two additional sub-networks to approximate the processing in Hui et al.'s algorithm~\cite{Hui_2018_CVPR}. However, once trained, we find it is optimal to directly use the reflectance chromaticity  estimates with the exact algorithm in \cite{Hui_2018_CVPR}, over the output of these sub-networks.

\subsection{Loss functions}
\paragraph{ChromNet loss.}
For the reflectance chromaticity estimation task, we use a scale-invariant loss. 
We also incorporate $\ell_1$ loss in gradient domain, to enforce that the estimated reflectance chromaticity is piece-wise constant. 
In particular, we define our loss function as
{{\small \begin{equation}
\mathcal{L}_{\alpha} = \frac{1}{M}\sum_{i = 1}^{M} \|\alpha^{*}_i - c_{\alpha}\alpha_i\|_1 +  \sum_{t = 1}^{L}\frac{1}{M_t}\sum_{i = 1}^{M_t} \| \nabla \alpha^{*}_{t, i} - c_{\alpha} \nabla \alpha_{t, i}  \|_1 ,
\label{eq:chrom_loss}
\end{equation}}}
where $\alpha^*$ denotes the predicted chromaticity, $\alpha$ is the ground truth provided, and  $c_\alpha$ is a term to compensate for the global scale difference, which can be estimated via least squares.  
We also use a mask to disregard the loss at pixels where we do not have reliable ground truth (e.g. pixels that are close to black or pixels corresponding to the outdoor environment map in the SUNCG dataset).
$M$ indicates the total number of valid pixels in an image.
Similar to the approach of Li et al.~\cite{li2018cgintrinsics}, we include a multi-scale matching term, where $L$ is the total number of layers specified ($3$ in the paper) and $M_t$ denotes the corresponding number of pixels not masked as invalid pixels.

\paragraph{ShadingNet loss.}
We impose an $\ell_2$ loss on both the absolute value and the gradients of the relative shadings. This encourages spatially smooth shading solutions (as is commonly done in prior intrinsic images work).
However, the network outputs two potential relative shadings and swapping these two predictions should not induce any loss.
To address this, we define our loss function as 
\[\mathcal{L}_{S} = \min\{\mathcal{L}_{S_{11}} + \mathcal{L}_{S_{22}}, \mathcal{L}_{S_{12}}+ \mathcal{L}_{S_{21}}\}\]
where $\mathcal{L}_{S_{ij}}$ denote the loss between the $i$-th output with $j$-th illuminant shadings defined in (\ref{eq:illuminant_shadings}).
Specifically, $\mathcal{L}_{S_{ij}}$ is defined as $\mathcal{L}_{S_{ij}} = \mathcal{L}_{\textrm{data}(i,j)} + \mathcal{L}_{\textrm{grad}(i,j)}$, where

{{\small
\begin{equation}
\mathcal{L}_{\textrm{data}(i,j)} = \frac{1}{M}\sum_{u = 1}^{M} \|S^{*}_{i, u} - c_SS_{j, u}\|_2,
\label{eq:shading_data_loss}
\end{equation}
\begin{equation}
\mathcal{L}_{\textrm{grad}(i,j)} = \sum_{t = 1}^{L}\frac{1}{M_t}\sum_{u = 1}^{M_t} \| \nabla S^{*}_{i, t, u} - c_S \nabla S_{i, t, j}  \|_2,
\label{eq:shading_grad_loss}
\end{equation}
}}
Here, $S^*_i$ denotes the $i$-th illuminant shading prediction while $S_j$ is the ground truth, and $c_S$ is the global scale to compensate for the illuminant brightness.  

\paragraph{SeparateNet loss.} 
Our loss for the two separated images is similar to our ShadingNet loss:
\[\mathcal{L}_{I} = \min\{\mathcal{L}_{I_{11}} + \mathcal{L}_{I_{22}}, \mathcal{L}_{I_{12}}+ \mathcal{L}_{I_{21}}\},\]
where $\mathcal{L}_{I_{ij}}$ is the $\ell_1$ loss.
Specifically, $\mathcal{L}_{I_{ij}}$ is defined as 
{{\small \begin{equation}
\mathcal{L}_{I_{ij}} =  \frac{1}{M}\sum_{u = 1}^{M} \|I^{*}_{i, u} - c_I I_{j, u}\|_1,
\label{eq:recon_loss}
\end{equation}}}
where, $I^*_i$ denote the $i$-th separated image predication while $I_j$ is the ground truth for the $j$-th light source, and $c_I$ is scale factor for the global intensity difference.

\paragraph{Training details.} 
We resize our training images to $384 \times 512$.
We use Adam optimizer ~\cite{kingma2014adam} to train our network with $\beta_1 = 0.5$. 
The initial learning rate is set to be $5\times 10^{-4}$
for all sub-networks. 
We cut down the learning rate by $1/10$ after $35$ epochs. 
We then train for $5$ epochs with the reduced learning rate. 
We ensure that all our networks have converged with this scheme.

\begin{table*}[!ttt]
	\centering
	\begin{tabular}{c | c | c} 
	\Xhline{4\arrayrulewidth}
	Name & Network architecture & Supervision \\
	\Xhline{4\arrayrulewidth}
	Chrom-Only & ChromNet & Chromaticity \\
	Final-Only & ChromNet + ShadingNet + SeparateNet &  Sep. images\\
	Full-Direct & ChromNet + ShadingNet + SeparateNet &  Chromaticity + Shadings + Sep. images\\
	SingleNet & Single Unet &  Sep. images\\
	\Xhline{4\arrayrulewidth}
	\end{tabular}
		\caption{Variant versions of proposed network architectures with different supervisions.}
		\label{table:baseline_names}
\end{table*}

\section{Evaluation}

We now present an extensive quantitative and qualitative evaluation of our proposed method.
Please refer to our supplementary material for more details and results.

\subsection{Test dataset}

\paragraph{Synthetic benchmark dataset.}
To quantitatively evaluate our method, we utilize the high quality synthetic dataset of ~\cite{bonneel2017intrinsic}.
This dataset has approximately $52$ scenes, each rendered under several different single illuminants. 
We first white balance each image of the same scene, and then modulate the white-balanced images with pre-selected light colors; these represent the ground truth separated images. 
The input images are then created by adding pairs of these separated images, each corresponding to one of the lights in the scene. 
We produce $400$ test samples in the dataset, and evaluate our method using both the ground truth of reflectance chromaticity and separated results.

\paragraph{Real dataset.}
We also evaluate the performance of our proposed technique on real images captured for both indoor and outdoor scenes.
Specifically, we utilize the dataset of the indoor scenes collected by Hui et al.~\cite{Hui_2018_CVPR} as well as  time-lapse videos for outdoor scenes.
Hui et al.~\cite{Hui_2018_CVPR} capture a pair of flash/no-flash for the same scene. 
We take the no-flash images in the dataset as the input to the network.
For the time-lapse videos, each frame serves as a test input as shown in Figure~\ref{fig:teaser} (a).

\paragraph{Error metric.} 
We characterize the performance of our approach on both reflectance chromaticity and the separated images.
We adopt the $\ell_1$ error to quantitative measure the performance of the reflectance chromaticity.
To evaluate the performance of the separated results, we compute the error for the separated result against the ground truth as:
\begin{equation}
\textrm{Loss} = \min\{E_{I_{1,1}} + E_{I_{2,2}}, E_{I_{1,2}} + E_{I_{2,1}}\}
\end{equation}
where $E$ denote the $\ell_1$ error between two images.
We use a global scale-invariant loss because we are most interested in capturing relative variations between the two images. %and errors in overall brightness.

%Specifically, we define the RMSE as:
%\begin{equation}
% \textrm{RMSE} = -10\cdot\log\left(\frac{\|cI^*_{i} - I_{i}\|_2}{\|I_{i}\|_2} \right),
% \label{eq:rmse}
%\end{equation}
%where $I^*$ and $I$ denote the estimated and ground truth images, respectively and $c$ denotes the scale in each color channel to compensate for the global difference.
\begin{table}[!ttt]
	\centering
	\begin{tabular}{r  | c c}  
		\Xhline{4\arrayrulewidth}
		Methods  & Chromaticity  & Separated Images   \\ 
		
		\Xhline{4\arrayrulewidth}
		\textbf{Proposed~~~~~~~~~~~}&&\\
		Chrom-Only  & \textbf{0.0308} & 0.0398  \\ [.5ex]
		Final-Only    & --- &0.0351   \\ [.5ex]
		Full-Direct  & 0.0537 & 0.0288 \\ [.5ex]
		Full+\cite{Hui_2018_CVPR}    & 0.0537 & \textbf{0.0207}   \\ [.5ex]
		\Xhline{2\arrayrulewidth}
		SingleNet  & --- &0.0679   \\ [.5ex]
		Shen et al.~\cite{shen2013intrinsic}  & 0.0821 & 0.0791  \\ [.5ex]
		Bell et al.~\cite{bell2014intrinsic}    & 0.0785 &0.0763   \\ [.5ex]
		Li et al.~\cite{li2018cgintrinsics}  & 0.0833 & 0.0821 \\ [.5ex]
		\Xhline{4\arrayrulewidth}
		$\dagger$Hsu et al.~\cite{hsu2008light}  & --- & 0.0678 \\ [.5ex]
		$\dagger$Hui et al.~\cite{Hui_2018_CVPR}  & --- & {0.0101} \\ [.5ex]
		%TwoLayerNet  & 0.00 & 0.00  \\ [.5ex]
		%Hui et al.~\cite{Hui_2018_CVPR} w. ground truth & --- & --- &--- &45.84\\[.5ex]
		\Xhline{3\arrayrulewidth}
		
	\end{tabular} 
	\vspace{2mm}
	{\small ~~~~~~~~~~~~~~~~~~~~~~~~~~~~~~~~~~~$\dagger$ Use additional information as input.}
	\caption{We measure performance of versions of our network---trained with different kinds of supervision, and with different approaches to perform separation---as well as other baselines. Reported here are $\ell_1$ error values for both estimated reflectance chromaticity (when available), as well as the final separated images.}
	\label{table:ablation_study}
\end{table}
\begin{comment}
Chromnet & \textbf{0.0253}  & 0.0166
DiectNet & --- & 0.0155
InDirectNet &---  & 0.0117
IndirectNet-sup (output) & --- & 0.0282
IndirectNet-sup (chromaticity) & 0.0366  & \textbf{0.0160}
\end{comment}
\begin{figure*}[!ttt]
	\small
	\centering
	\setlength{\tabcolsep}{1pt}
	\begin{tabular}{ccccc}
		\includegraphics[width=0.16\textwidth]{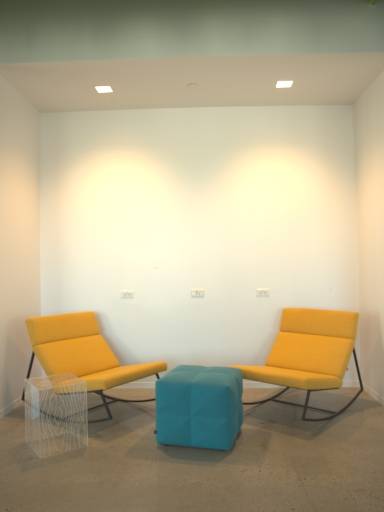}& 
		\includegraphics[width=0.16\textwidth]{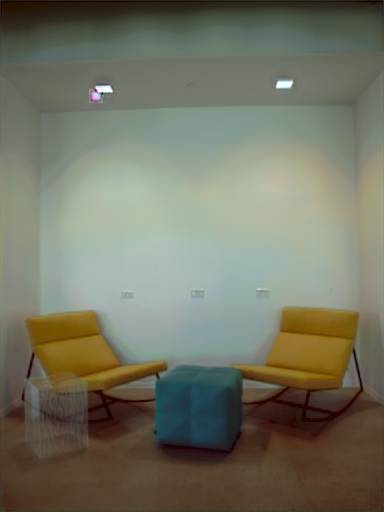}& 
		\includegraphics[width=0.16\textwidth]{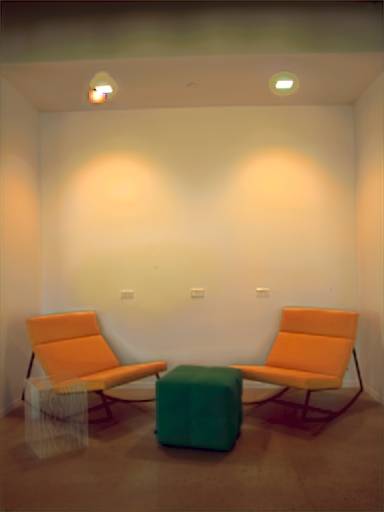} &
		\includegraphics[width=0.16\textwidth]{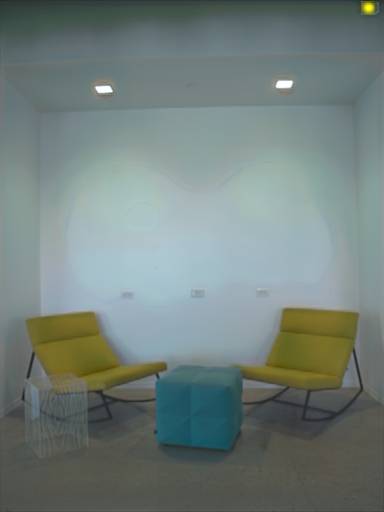} &
		\includegraphics[width=0.16\textwidth]{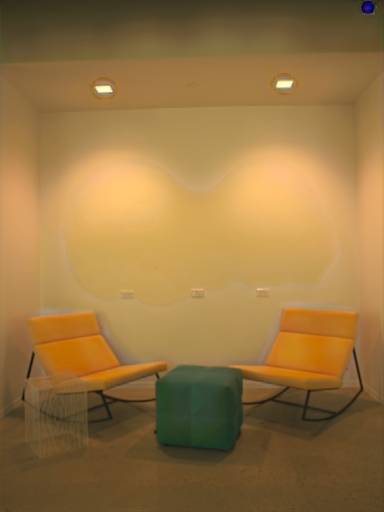} \\
		(a) Input & \multicolumn{2}{c}{(b) SingleNet} & \multicolumn{2}{c}{(c) Final-Only} \\
		
	\end{tabular}
	\begin{tabular}{cccccc}
		\includegraphics[width=0.16\textwidth]{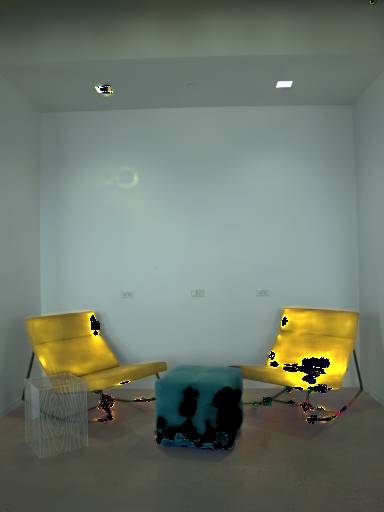} &
		\includegraphics[width=0.16\textwidth]{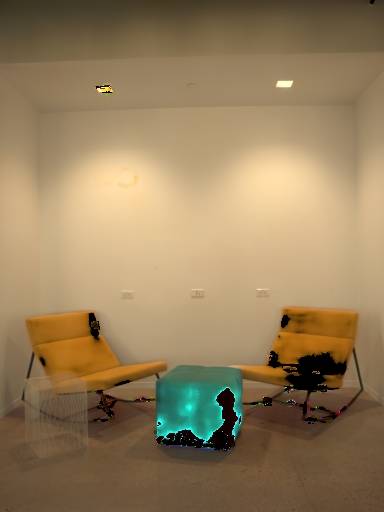} &
		\includegraphics[width=0.16\textwidth]{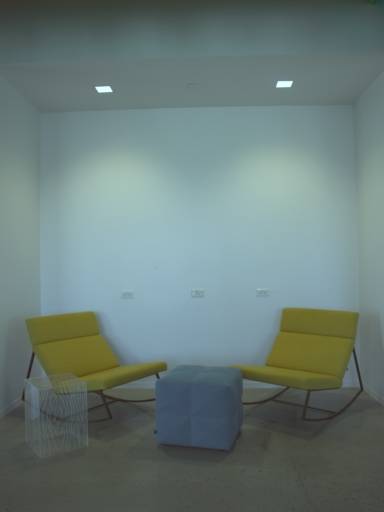}& 
		\includegraphics[width=0.16\textwidth]{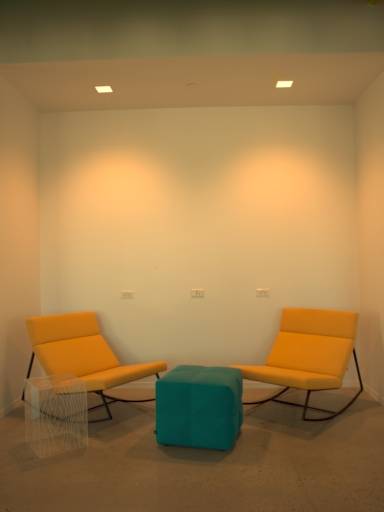}& 
		\includegraphics[width=0.16\textwidth]{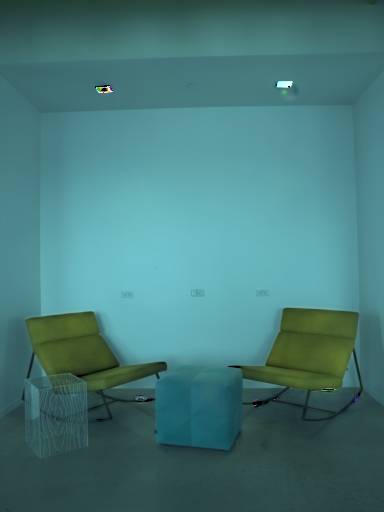} &
		\includegraphics[width=0.16\textwidth]{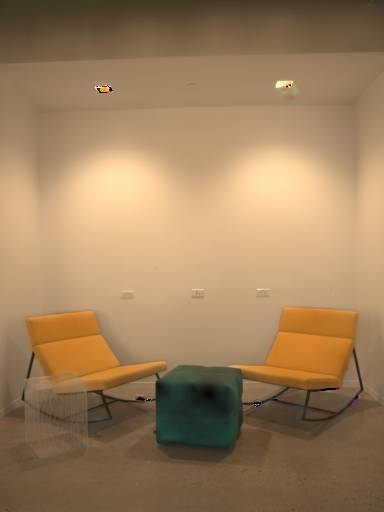} \\
		\multicolumn{2}{c}{(d) Chrom-Only} & \multicolumn{2}{c}{(e) Full-Direct} & \multicolumn{2}{c}{(f) Full+\cite{Hui_2018_CVPR}}
	\end{tabular}
	
	\caption{Qualitative comparison of image separation results  of different versions of our network, as well as of the single encoder-decoder architecture network (SingleNet). We see that both SingleNet and our Final-only model both fail to separate the effects of illuminant shading from the input. Our Chrom-Only model yields a better result, but has severe artifacts in certain regions---highlighting that better chromaticity estimates do not lead to better separation. The results from our model with Full supervision yields the best results---with better separation of shadow and shading effects when we use its chromaticity outputs in conjunction with \cite{Hui_2018_CVPR}.}	
	\label{fig:comp_real}
\end{figure*} 

\subsection{Quantitative results on synthetic benchmark}

In Table~\ref{table:ablation_study}, we report the performance of our approach, and compare it to several baselines (summarized in Table~\ref{table:baseline_names}). We begin by quantifying the importance of supervision. We train different models for our network: with full supervision, with supervision only on the quality of the final separated images (\textbf{Final-Only}), and training only the first sub-network, i.e., ChromNet, with supervision only on reflectance chromaticities (\textbf{Chrom-Only}). Moreover, for our fully supervised model (\textbf{Full}), we consider using the separated images directly predicted by our full network (\textbf{Full-Direct}), as well as taking only the reflectance chromaticity estimates and using Hui et al.'s algorithm~\cite{Hui_2018_CVPR}---which includes more complex processing---to perform the separation (\textbf{Full+\cite{Hui_2018_CVPR}}). For the model with only chromaticity supervision, we also use \cite{Hui_2018_CVPR} perform separation, and for the final-only supervised model (where intermediate chromaticities are not meaningful), we only consider the final output.

We find that our model trained with full supervision has the best performance in terms of the quality of final separated images. Interestingly, the \textbf{Chrom-Only} model is better at predicting chromaticity, but as expected, this does not translate to higher quality image outputs. The \textbf{Final-Only} model also yields worse separation results despite being trained with respect to their quality, highlighting the importance of intermediate supervision. Finally, we find that using our \textbf{Full} model in combination with \cite{Hui_2018_CVPR} yields comparatively better results than taking the direct final output of the network. Thus, our final sub-networks (ShadingNet and SeparateNet) are able to only approximate \cite{Hui_2018_CVPR}'s algorithm. Thus, their main benefit in our framework is in allowing back-propagation to provide supervision for chromaticity estimation, in a manner that is optimal for separation.

We also include comparisons to a network with a more traditional architecture (rather than three sub-networks) to do direct separation (\textbf{SingleNet}). We use the same architecture as the encoder-decoder portion of our ShadingNet, and train this again with supervision only on the final separated outputs. We find that this performs significantly worse (than even \textbf{Final-Only}), illustrating the utility of our physically-motivated architecture. Finally, we also include the comparisons with baselines where different intrinsic image decomposition methods~\cite{shen2013intrinsic,bell2014intrinsic,li2018cgintrinsics} are used to estimate reflectance chromaticity from a single image, and these are used for separation with \cite{Hui_2018_CVPR}. We find these methods yield lower accuracy in both reflectance chromaticity estimation and lighting separation---likely because they, like most intrinsic image methods, assume a single light source.

Finally, we evaluate on two methods that require additional information beyond a single image: ground truth light colors for Hsu et al.~\cite{hsu2008light}, and a flash/no-flash pair which provides direct access to reflectance chromaticity, for Hui et al.~\cite{Hui_2018_CVPR}. We produce better results than \cite{hsu2008light}, but as expected,  \cite{Hui_2018_CVPR} yields the most accurate separation, since it has direct access to chromaticity information---but requires capturing an additional flash image.

\begin{figure*}[!ttt]
	{
		\small
		\centering
		\setlength{\tabcolsep}{1pt}		
		\begin{tabular}{ccc}
			\includegraphics[width=0.25\textwidth]{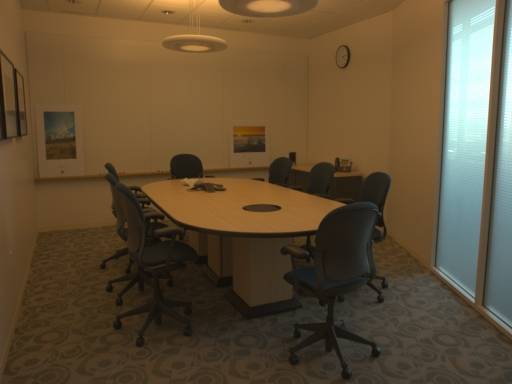} &
			\includegraphics[width=0.25\textwidth]{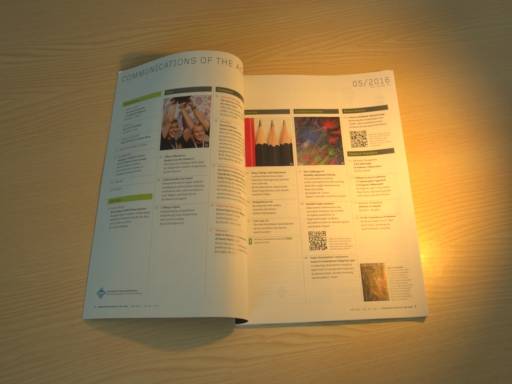} &
			\includegraphics[width=0.25\textwidth]{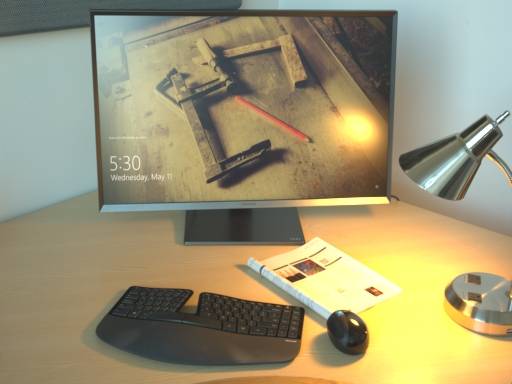} \\
			\multicolumn{3}{c}{(a) Input photographs} 
\end{tabular}
\begin{tabular}{cccc}

\includegraphics[width=0.25\textwidth]{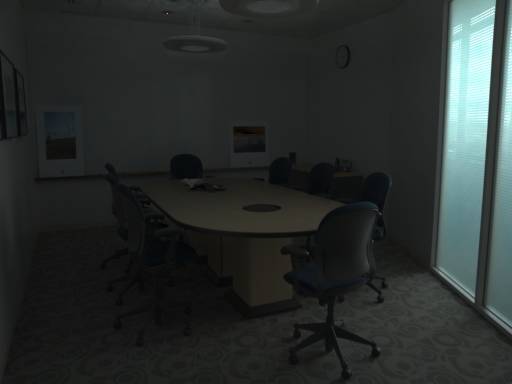}&
\includegraphics[width=0.25\textwidth]{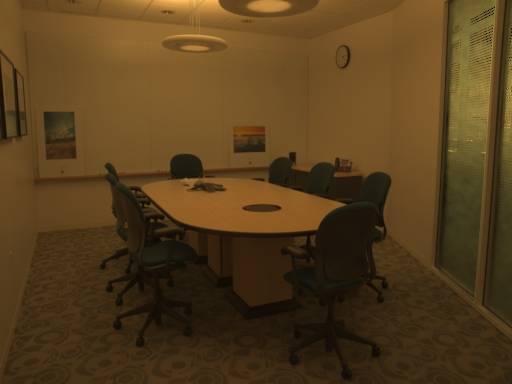}&
\includegraphics[width=0.25\textwidth]{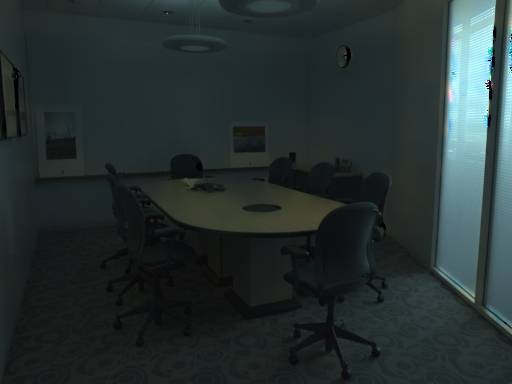}&
\includegraphics[width=0.25\textwidth]{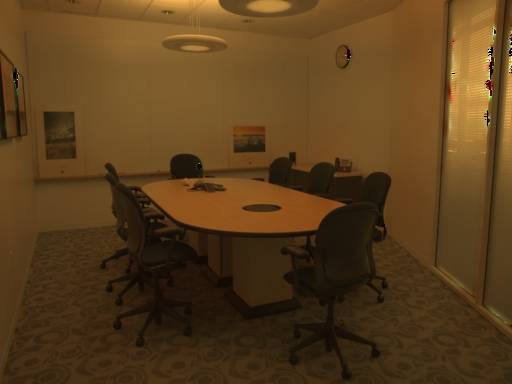}\\

\includegraphics[width=0.25\textwidth]{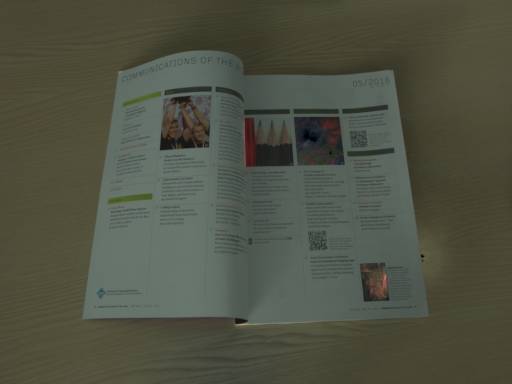}&
\includegraphics[width=0.25\textwidth]{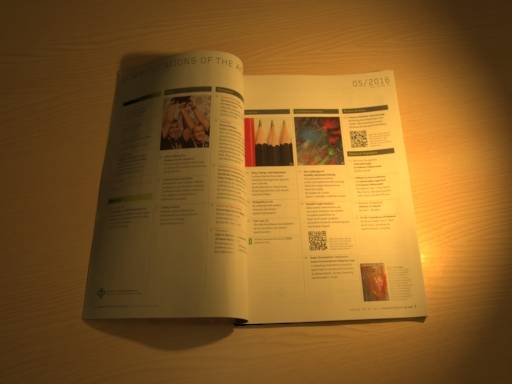}&
\includegraphics[width=0.25\textwidth]{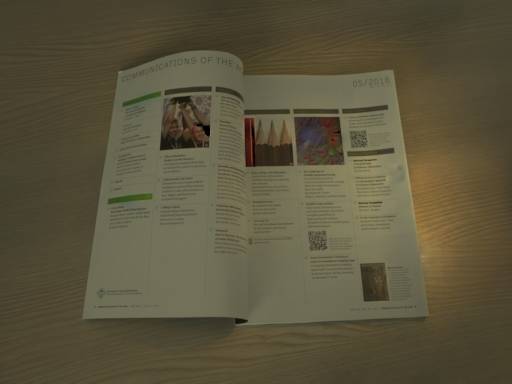}&
\includegraphics[width=0.25\textwidth]{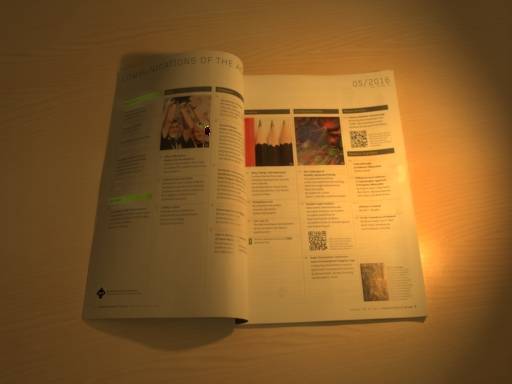}\\%
\includegraphics[width=0.25\textwidth]{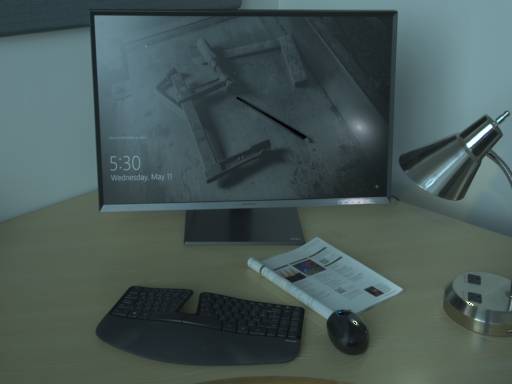}&
\includegraphics[width=0.25\textwidth]{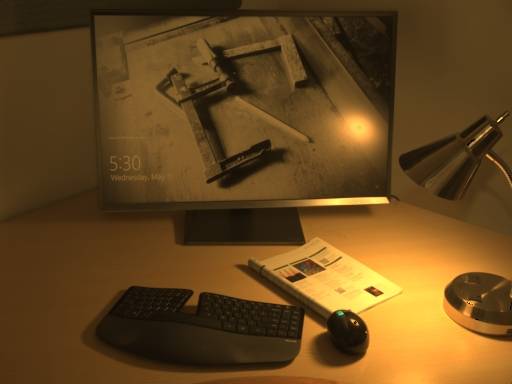}&
\includegraphics[width=0.25\textwidth]{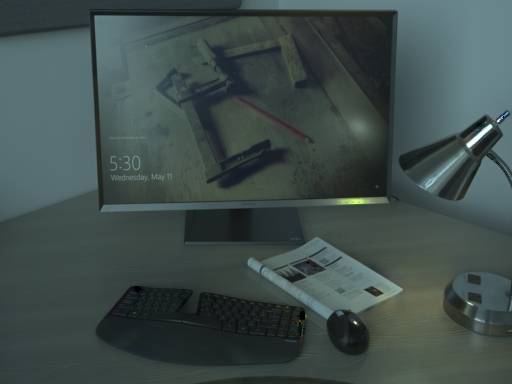}&
\includegraphics[width=0.25\textwidth]{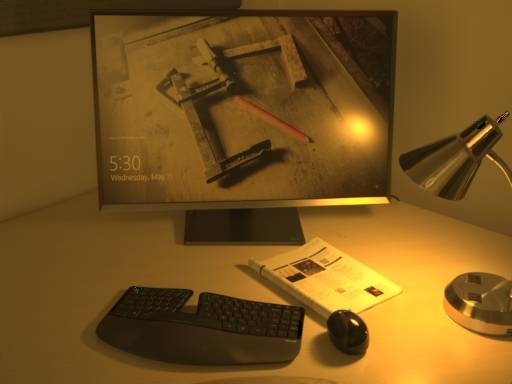}\\		
\multicolumn{2}{c}{(b) Hui et al.~\cite{Hui_2018_CVPR}} &
\multicolumn{2}{c}{(c) Ours} \\
\end{tabular}
		
\caption{ We evaluate our technique against the flash photography technique by Hui et al~\cite{Hui_2018_CVPR}. While the proposed method may lead to small artifacts in the resulting image, we can achieve nearly the same visual quality as Hui et al.~\cite{Hui_2018_CVPR}, which captures two photographs for the same scene. In comparison, the proposed technique by using single photograph yields more practical solution to the problem.
		}
		\label{fig:comp_hui}
		
	}
\end{figure*}

\begin{figure*}[!ttt]
{
	\small
	\centering
	\setlength{\tabcolsep}{1pt}		
\begin{tabular}{ccccc}
\includegraphics[width=0.2\textwidth]{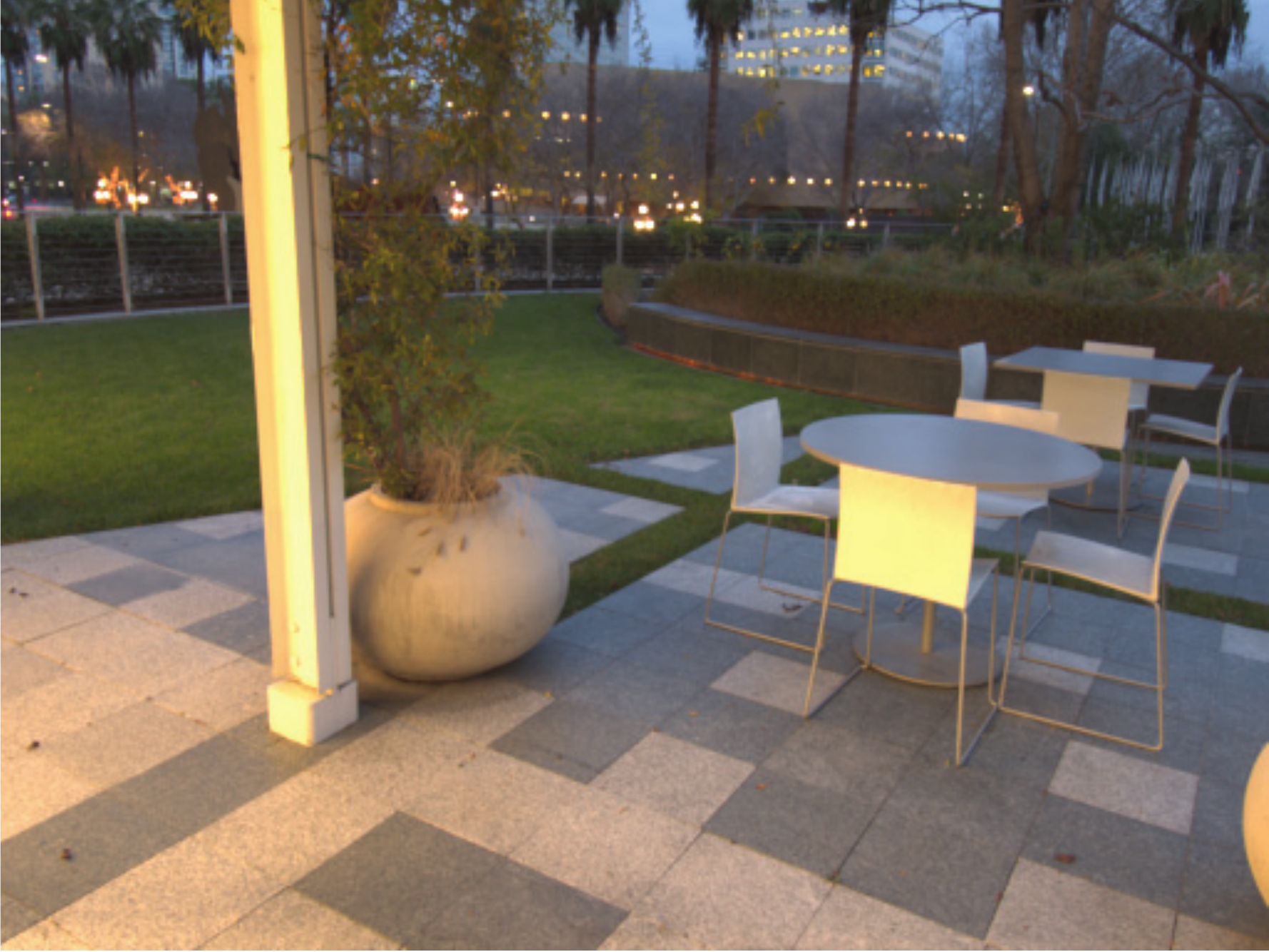}&
\includegraphics[width=0.2\textwidth]{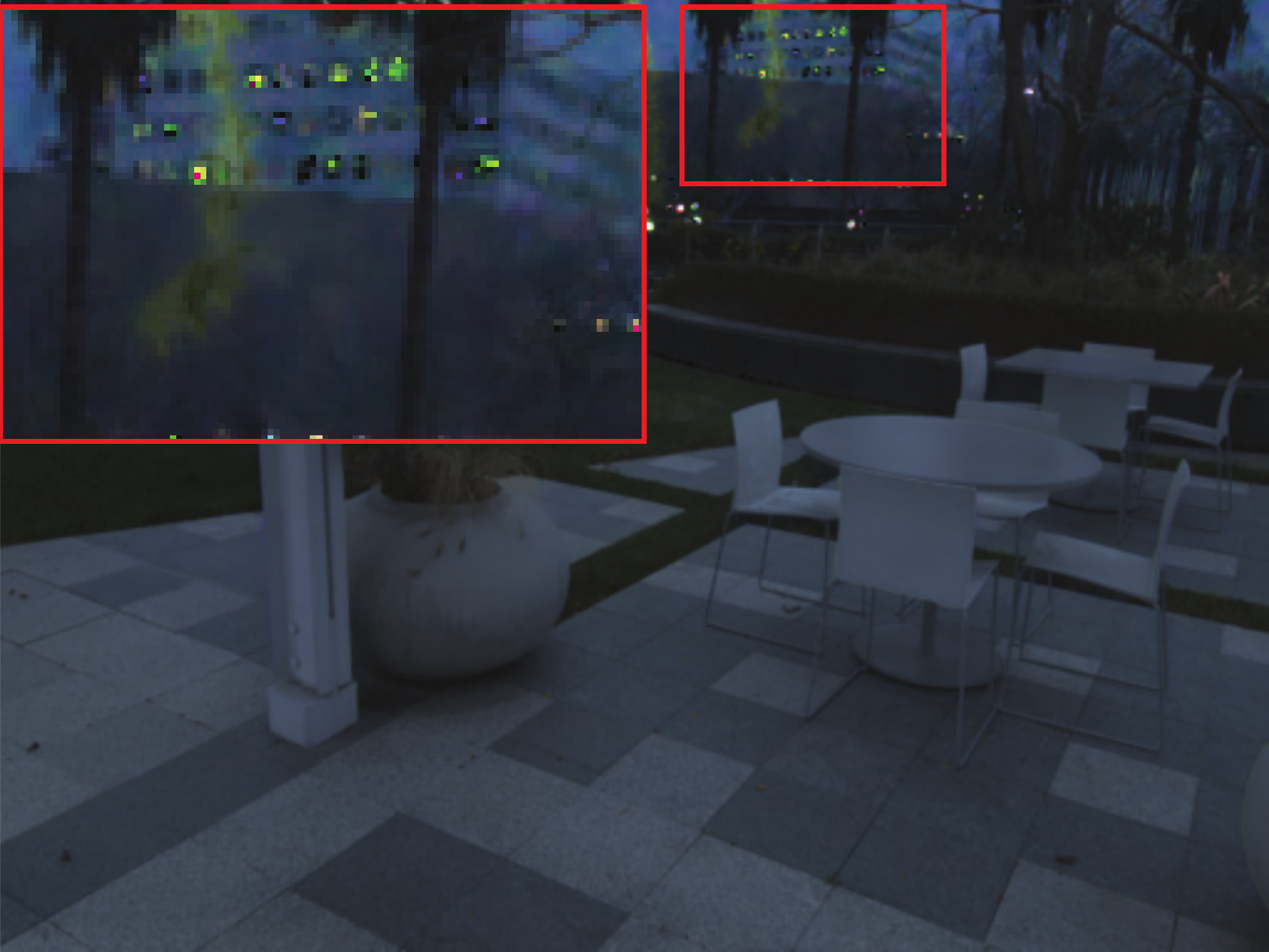}&
\includegraphics[width=0.2\textwidth]{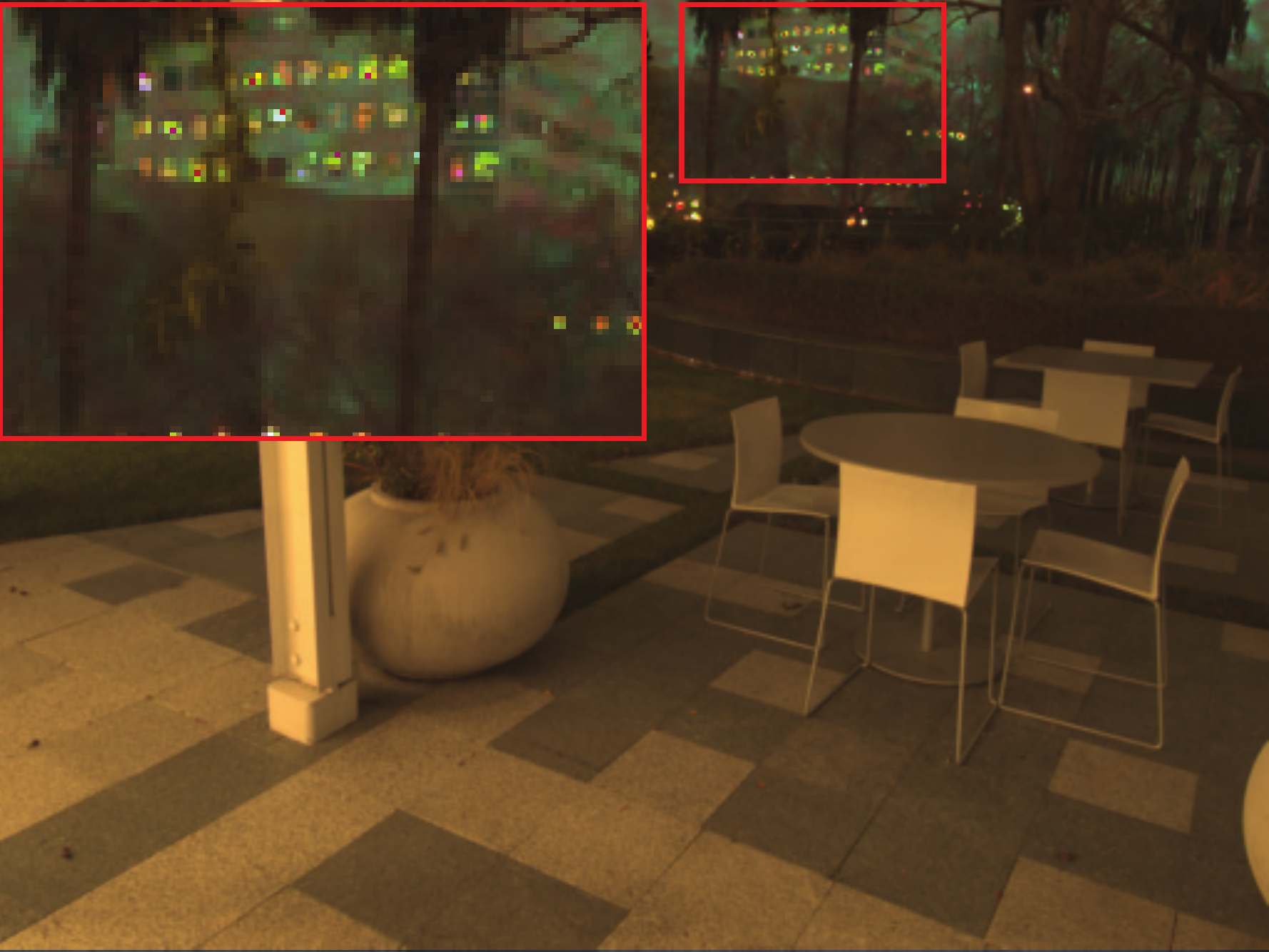}&
\includegraphics[width=0.2\textwidth]{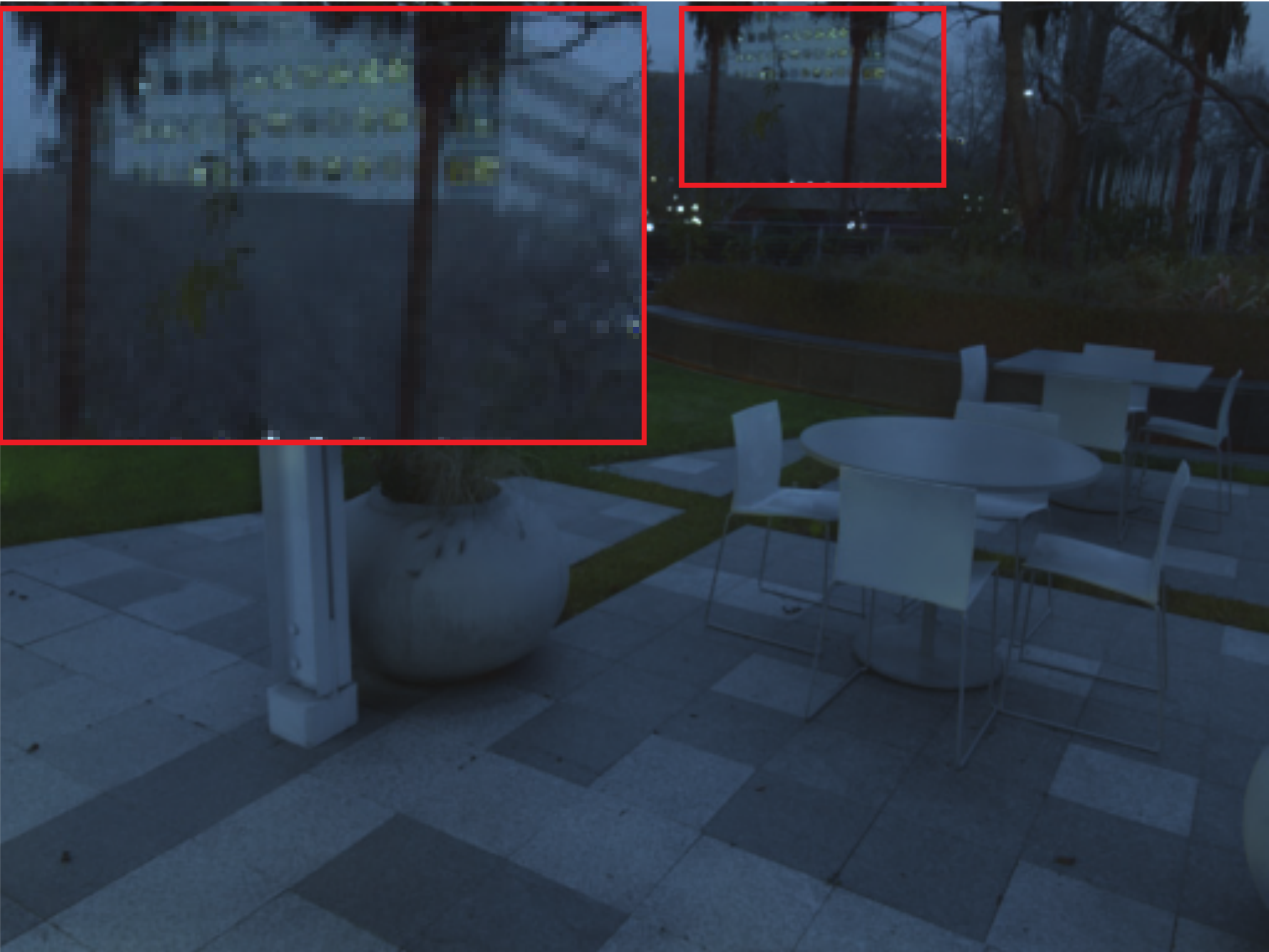}&
\includegraphics[width=0.2\textwidth]{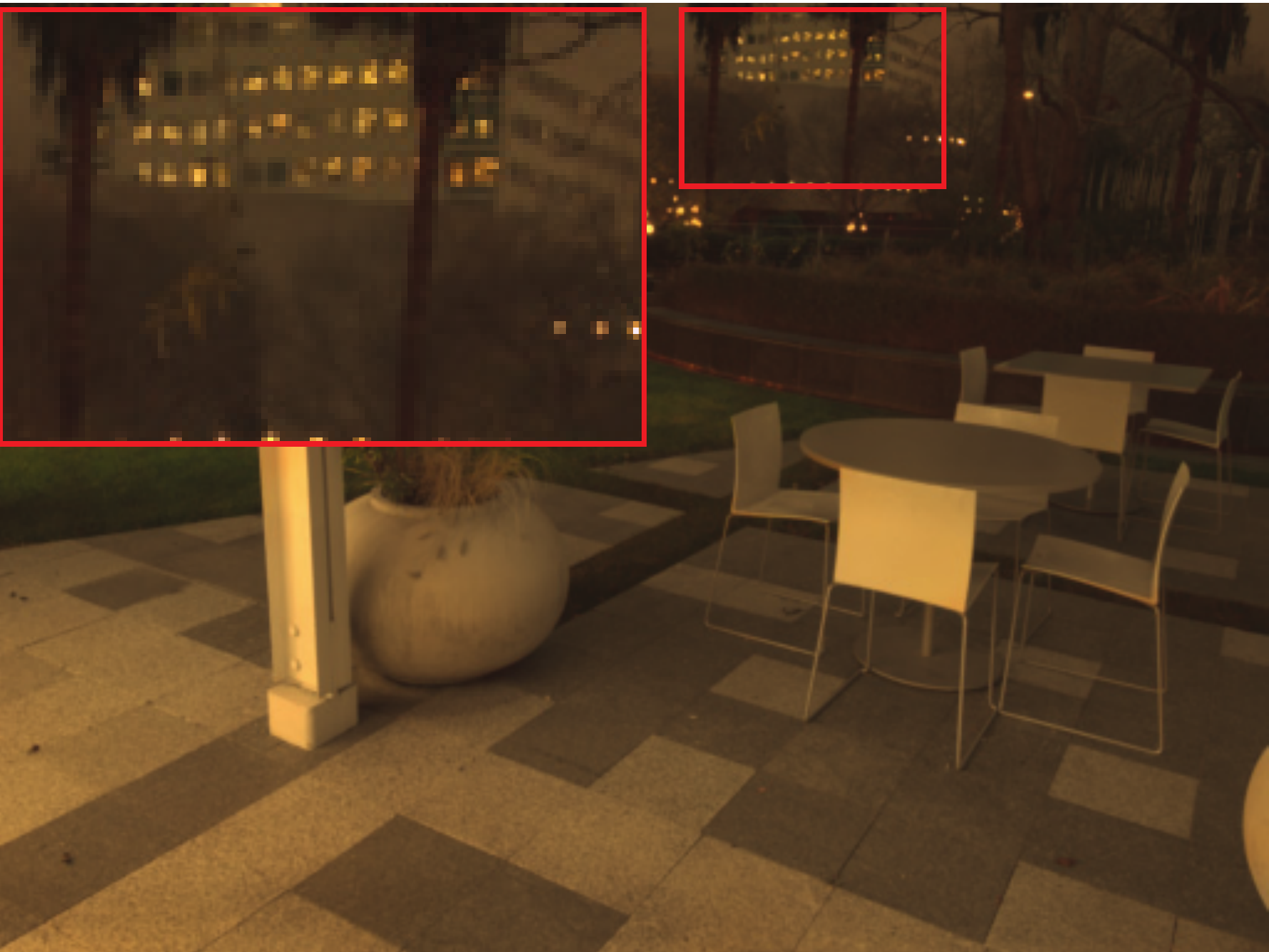}\\	
(a) Input &\multicolumn{2}{c}{(b) Hui et al.~\cite{Hui_2018_CVPR}} &
\multicolumn{2}{c}{(c) Our} \\
\end{tabular}
		
\caption{For the outdoor scene (a), the flash is not strong enough to illuminate the far-away scene points, which results in the artifacts in (b). In contrast, our method takes a single photograph and does not rely on flash illumination. As can be seen, the artifacts can be eliminated and visual quality has been significantly improved.
		}
\label{fig:out_door}
		
	}
\end{figure*}

\subsection{Qualitative evaluation on real data}
Figure~\ref{fig:comp_real} shows results on a real image for the different versions of our network (as well as of \textbf{SingleNet}), while Figure~\ref{fig:comp_hui} compares our results comparison to Hui et al.'s method~\cite{Hui_2018_CVPR} when using a flash/no-flash pair. These results confirm our conclusions from Table~\ref{table:ablation_study}---we find that the version of our network trained with full supervision performs best, especially when used in combination with \cite{Hui_2018_CVPR} to carry out the separation from predicted chromaticities. Moreover, despite requiring only a single image input, it comes close to matching Hui et al.'s~\cite{Hui_2018_CVPR} performance with a flash/no-flash pair. We show an example in Figure~\ref{fig:out_door} where our method affords a distinct advantage even when an image with flash is available, but when several regions in the scene are too far from the flash. This leads to artifacts in those regions for \cite{Hui_2018_CVPR}, while our approach is able to perform a higher quality separation.

\section{Conclusions}
We describe a learning-based approach to separate the lighting effect of two illuminants in an image. Our method relies on the use of a deep-neural network based estimator, whose architecture is motivated by a physics-based analysis of the problem and associated intermediate supervision. Our ablation experiments demonstrate the importance of this supervision. Crucially, we show that we are able to produce high-quality outputs that match the performance of previous methods that required a flash/no-flash pair, while being more practical in requiring only a single image.

\section{Acknowledgment}
This research was supported, in part, by the National Geospatial-Intelligence Agency's Academic Research
Program (Award No. HM0476-17-1-2000), the NSF CAREER grant CCF-1652569, and a gift from Adobe Research. Chakrabarti was supported by the NSF Grant IIS-1820693.


{\small
\begin{thebibliography}{10}\itemsep=-1pt
	
\bibitem{flashambient}
Ya\u{g}{\i}z Aksoy, Changil Kim, Petr Kellnhofer, Sylvain Paris, Mohamed
  Elgharib, Marc Pollefeys, and Wojciech Matusik.
\newblock A dataset of flash and ambient illumination pairs from the crowd.
\newblock In {\em ECCV}, 2018.

\bibitem{barron2012color}
Jonathan~T Barron and Jitendra Malik.
\newblock Color constancy, intrinsic images, and shape estimation.
\newblock In {\em ECCV}. 2012.

\bibitem{barron2015shape}
Jonathan~T Barron and Jitendra Malik.
\newblock Shape, illumination, and reflectance from shading.
\newblock {\em PAMI}, 37(8):1670--1687, 2015.

\bibitem{Barrow78:Intrinsic}
H. Barrow and J. Tenenbaum.
\newblock Recovering intrinsic scene characteristics from images.
\newblock {\em Computer Vision Systems}, 1978.

\bibitem{bell2014intrinsic}
Sean Bell, Kavita Bala, and Noah Snavely.
\newblock Intrinsic images in the wild.
\newblock {\em TOG}, 33(4):159, 2014.

\bibitem{bonneel2017intrinsic}
Nicolas Bonneel, Balazs Kovacs, Sylvain Paris, and Kavita Bala.
\newblock Intrinsic decompositions for image editing.
\newblock In {\em Computer Graphics Forum}, 2017.

\bibitem{bonneel2014interactive}
Nicolas Bonneel, Kalyan Sunkavalli, James Tompkin, Deqing Sun, Sylvain Paris,
  and Hanspeter Pfister.
\newblock Interactive intrinsic video editing.
\newblock {\em TOG}, 33(6):197, 2014.

\bibitem{bousseau2009user}
Adrien Bousseau, Sylvain Paris, and Fr{\'e}do Durand.
\newblock User-assisted intrinsic images.
\newblock In {\em TOG}, volume~28, page 130, 2009.

\bibitem{boyadzhiev2012user}
Ivaylo Boyadzhiev, Kavita Bala, Sylvain Paris, and Fr{\'e}do Durand.
\newblock User-guided white balance for mixed lighting conditions.
\newblock {\em TOG}, 31(6):200, 2012.

\bibitem{boyadzhiev2013user}
Ivaylo Boyadzhiev, Sylvain Paris, and Kavita Bala.
\newblock User-assisted image compositing for photographic lighting.
\newblock {\em TOG}, 32(4):36--1, 2013.

\bibitem{debevec2008rendering}
Paul Debevec.
\newblock Rendering synthetic objects into real scenes: Bridging traditional
  and image-based graphics with global illumination and high dynamic range
  photography.
\newblock In {\em SIGGRAPH 2008 classes}, page~32, 2008.

\bibitem{Debevec12:LightStage}
Paul Debevec.
\newblock The {Light} {Stages} and {Their} {Applications} to {Photoreal}
  {Digital} {Actors}.
\newblock In {\em {SIGGRAPH} {Asia}}, 2012.

\bibitem{ebner2004color}
Marc Ebner.
\newblock Color constancy using local color shifts.
\newblock In {\em ECCV}, 2004.

\bibitem{finlayson1993enhancing}
Graham Finlayson, MS Drew, and BV Funt.
\newblock Enhancing von kries adaptation via sensor transformations.
\newblock 1993.

\bibitem{finlayson1993diagonal}
Graham~D Finlayson, Mark~S Drew, and Brian~V Funt.
\newblock Diagonal transforms suffice for color constancy.
\newblock In {\em ICCV}, 1993.

\bibitem{gardner2017indoor}
Marc-Andre Gardner, Kalyan Sunkavalli, Ersin Yumer, Xiaohui Shen, Emiliano
  Gambaretto, Christian Gagné, and Jean-François Lalonde.
\newblock Learning to predict indoor illumination from a single image.
\newblock {\em TOG}, 9(4), 2017.

\bibitem{gehler2008bayesian}
Peter~Vincent Gehler, Carsten Rother, Andrew Blake, Tom Minka, and Toby Sharp.
\newblock Bayesian color constancy revisited.
\newblock In {\em CVPR}, 2008.

\bibitem{Gijsenij2011survey}
A. Gijsenij, T. Gevers, and J. van~de Weijer.
\newblock Computational color constancy: Survey and experiments.
\newblock {\em TIP}, 20(9):2475--2489, 2011.

\bibitem{gijsenij2012color}
Arjan Gijsenij, Rui Lu, and Theo Gevers.
\newblock Color constancy for multiple light sources.
\newblock {\em TIP}, 21(2):697--707, 2012.

\bibitem{grosse2009ground}
Roger Grosse, Micah~K Johnson, Edward~H Adelson, and William~T Freeman.
\newblock Ground truth dataset and baseline evaluations for intrinsic image
  algorithms.
\newblock In {\em CVPR}, 2009.

\bibitem{hauagge2013photometric}
Daniel Hauagge, Scott Wehrwein, Kavita Bala, and Noah Snavely.
\newblock Photometric ambient occlusion.
\newblock In {\em CVPR}, 2013.

\bibitem{hold2017deep}
Yannick Hold-Geoffroy, Kalyan Sunkavalli, Sunil Hadap, Emiliano Gambaretto, and
  Jean-Fran{\c{c}}ois Lalonde.
\newblock Deep outdoor illumination estimation.
\newblock In {\em CVPR}, 2017.

\bibitem{hsu2008light}
Eugene Hsu, Tom Mertens, Sylvain Paris, Shai Avidan, and Fredo Durand.
\newblock Light mixture estimation for spatially varying white balance.
\newblock In {\em TOG}, volume~27, page~70, 2008.

\bibitem{hui2016white}
Zhuo Hui, Aswin~C. Sankaranarayanan, Kalyan Sunkavalli, and Sunil Hadap.
\newblock White balance under mixed illumination using flash photography.
\newblock In {\em ICCP}, 2016.

\bibitem{Hui_2018_CVPR}
Zhuo Hui, Kalyan Sunkavalli, Sunil Hadap, and Aswin~C. Sankaranarayanan.
\newblock Illuminant spectra-based source separation using flash photography.
\newblock In {\em CVPR}, 2018.

\bibitem{Hui_2017_ICCV}
Zhuo Hui, Kalyan Sunkavalli, Joon-Young Lee, Sunil Hadap, Jian Wang, and
  Aswin~C. Sankaranarayanan.
\newblock Reflectance capture using univariate sampling of brdfs.
\newblock In {\em ICCV}, 2017.

\bibitem{isola2017image}
Phillip Isola, Jun-Yan Zhu, Tinghui Zhou, and Alexei~A Efros.
\newblock Image-to-image translation with conditional adversarial networks.
\newblock {\em CVPR}, 2017.

\bibitem{johnson2016perceptual}
Justin Johnson, Alexandre Alahi, and Li Fei-Fei.
\newblock Perceptual losses for real-time style transfer and super-resolution.
\newblock In {\em ECCV}, 2016.

\bibitem{kingma2014adam}
Diederik~P Kingma and Jimmy Ba.
\newblock Adam: A method for stochastic optimization.
\newblock {\em arXiv preprint arXiv:1412.6980}, 2014.

\bibitem{Laffont:ICCV15}
Pierre-Yves Laffont and Jean-Charles Bazin.
\newblock Intrinsic decomposition of image sequences from local temporal
  variations.
\newblock In {\em ICCV}, 2015.

\bibitem{lalonde2010sun}
Jean-Fran{\c{c}}ois Lalonde, Srinivasa~G Narasimhan, and Alexei~A Efros.
\newblock What do the sun and the sky tell us about the camera?
\newblock {\em IJCV}, 88(1):24--51, 2010.

\bibitem{li2018cgintrinsics}
Zhengqi Li and Noah Snavely.
\newblock Cgintrinsics: Better intrinsic image decomposition through
  physically-based rendering.
\newblock In {\em ECCV}, 2018.

\bibitem{li2018learning}
Zhengqi Li and Noah Snavely.
\newblock Learning intrinsic image decomposition from watching the world.
\newblock In {\em CVPR}, 2018.

\bibitem{li2018materials}
Zhengqin Li, Kalyan Sunkavalli, and Manmohan Chandraker.
\newblock Materials for masses: Svbrdf acquisition with a single mobile phone
  image.
\newblock In {\em ECCV}, 2018.

\bibitem{prinet2013illuminant}
Veronique Prinet, Dani Lischinski, and Michael Werman.
\newblock Illuminant chromaticity from image sequences.
\newblock In {\em ICCV}, 2013.

\bibitem{rematas2016deep}
Konstantinos Rematas, Tobias Ritschel, Mario Fritz, Efstratios Gavves, and
  Tinne Tuytelaars.
\newblock Deep reflectance maps.
\newblock In {\em CVPR}, 2016.

\bibitem{riess2011illuminant}
Christian Riess, Eva Eibenberger, and Elli Angelopoulou.
\newblock Illuminant color estimation for real-world mixed-illuminant scenes.
\newblock In {\em ICCVW}, 2011.

\bibitem{shen2013intrinsic}
Jianbing Shen, Xiaoshan Yang, Xuelong Li, and Yunde Jia.
\newblock Intrinsic image decomposition using optimization and user scribbles.
\newblock {\em IEEE Transactions on Cybernetics}, 43(2):425--436, 2013.

\bibitem{song2016ssc}
Shuran Song, Fisher Yu, Andy Zeng, Angel~X Chang, Manolis Savva, and Thomas
  Funkhouser.
\newblock Semantic scene completion from a single depth image.
\newblock {\em CVPR}, 2017.

\bibitem{sunkavalli08color}
Kalyan Sunkavalli, Fabiano Romeiro, Wojciech Matusik, Todd Zickler, and
  Hanspeter Pfister.
\newblock What do color changes reveal about an outdoor scene?
\newblock In {\em CVPR}, 2008.

\bibitem{tang2012deep}
Yichuan Tang, Ruslan Salakhutdinov, and Geoffrey Hinton.
\newblock Deep lambertian networks.
\newblock {\em arXiv preprint arXiv:1206.6445}, 2012.

\bibitem{zhao2012closed}
Qi Zhao, Ping Tan, Qiang Dai, Li Shen, Enhua Wu, and Stephen Lin.
\newblock A closed-form solution to retinex with nonlocal texture constraints.
\newblock {\em PAMI}, 34(7):1437--1444, 2012.

\bibitem{zhou2015learning}
Tinghui Zhou, Philipp Krahenbuhl, and Alexei~A Efros.
\newblock Learning data-driven reflectance priors for intrinsic image
  decomposition.
\newblock In {\em ICCV}, 2015.
	
\end{thebibliography}
}
\end{document}